\documentclass[runningheads]{llncs}
\usepackage[T1]{fontenc}
\usepackage[marginal]{footmisc}
\usepackage{amsmath,graphicx}
\usepackage{algpseudocode}
\usepackage{algcompatible}
\usepackage{arydshln}
\usepackage{adjustbox}
\usepackage{multirow} 
\usepackage{array}
\usepackage{tabularx}
\usepackage{booktabs}
\usepackage{algorithm} 
\usepackage{wrapfig}
\usepackage{tabu}                     
\usepackage{multirow}                 
\usepackage{multicol}                 
\usepackage{multirow}                
\usepackage{float}                    
\usepackage{makecell}                 
\usepackage{booktabs}                 
\usepackage[marginal]{footmisc}
\usepackage{marvosym}

\usepackage{pifont}
\titlerunning{T$^\text{3}$SVFND}

\begin{document}
\title{T$^\text{3}$SVFND: Towards an Evolving Fake News Detector for Emergencies with Test-time Training on Short Video Platforms}

\author{Liyuan Zhang
\and
Zeyun Cheng
\and Zhongyan Gui\textsuperscript{(\Letter)}  
\and
Yan Yang
\textsuperscript{(\Letter)}
\and
Yong Liu
\and
Jinke Ma
}
\authorrunning{Liyuan Zhang et al.}
\institute{Heilongjiang University, Harbin, China
\email {\{2231976,2231975\}@s.hlju.edu.cn,\{guizhongyan,yangyan,liuyong123456\}@hlju.edu.cn}
}
\renewcommand{\thefootnote}{}
\footnotetext{This work was supported by the National Natural Science Foundation of China (No. 6247074060), the Natural Science Foundation of Heilongjiang Province in China (No. PL2024F029) and the Basic Research Funds for Provincial Universities in Heilongjiang Province (No. 2023-KYYWF-1486, No. 2024-KYYWF-0115).}
\maketitle    
%

%

%
    
%
\begin{abstract}
The existing methods for fake news videos detection may not be generalized, because there is a distribution shift between short video news of different events, and the performance of such techniques greatly drops if news records are coming from emergencies. We propose a new fake news videos detection framework (T$^3$SVFND) using Test-Time Training (TTT) to alleviate this limitation, enhancing the robustness of fake news videos detection. Specifically, we design a self-supervised auxiliary task based on Mask Language Modeling (MLM) that masks a certain percentage of words in text and predicts these masked words by combining contextual information from different modalities (audio and video). In the test-time training phase, the model adapts to the distribution of test data through auxiliary tasks. Extensive experiments on the public benchmark demonstrate the effectiveness of the proposed model, especially for the detection of emergency news.
\end{abstract}
\section{Introduction}
\label{sec:intro}
Due to the characteristics of short duration, concentrated content and strong expression, sharing short videos have penetrated deeply into the daily life of the public. The low entry threshold, few self-censorship mechanisms and other factors have caused a large number of fake news on short video platforms. Previous image text fake news detection methods \cite{ma2022curriculum,singhal2019spotfake,zhou2023multi,zhou2023multimodal} are difficult to apply directly to news videos due to different modalities. Developing reliable methods for automated detection of fake news videos using artificial intelligence technology is currently a top priority.

Recently, a large-scale fake news short videos dataset (FakeSV) \cite{qi2023fakesv} has been proposed, a number of fake news videos detection methods have demonstrated their effectiveness on this dataset. Despite the success of these methods, when they are faced with emergency news, there is a generally sharp decline in performance, calls into question about their reliability. Traditional fake news detection models assume the same training and testing distribution, but actual deployment requires the capacity of model to generalize unseen and out-of-distribution data which can ensure that the effectiveness still remains in the dynamic real world. However, existing fake news videos detection methods \cite{hou2019towards,shang2021multimodal,serrano2020nlp,choi2021using} do not take this into account, which results in failing to maintain claimed performance on previously unseen events. 

\begin{wrapfigure}{r}{0.5\textwidth}
    \includegraphics[width=0.5\textwidth]{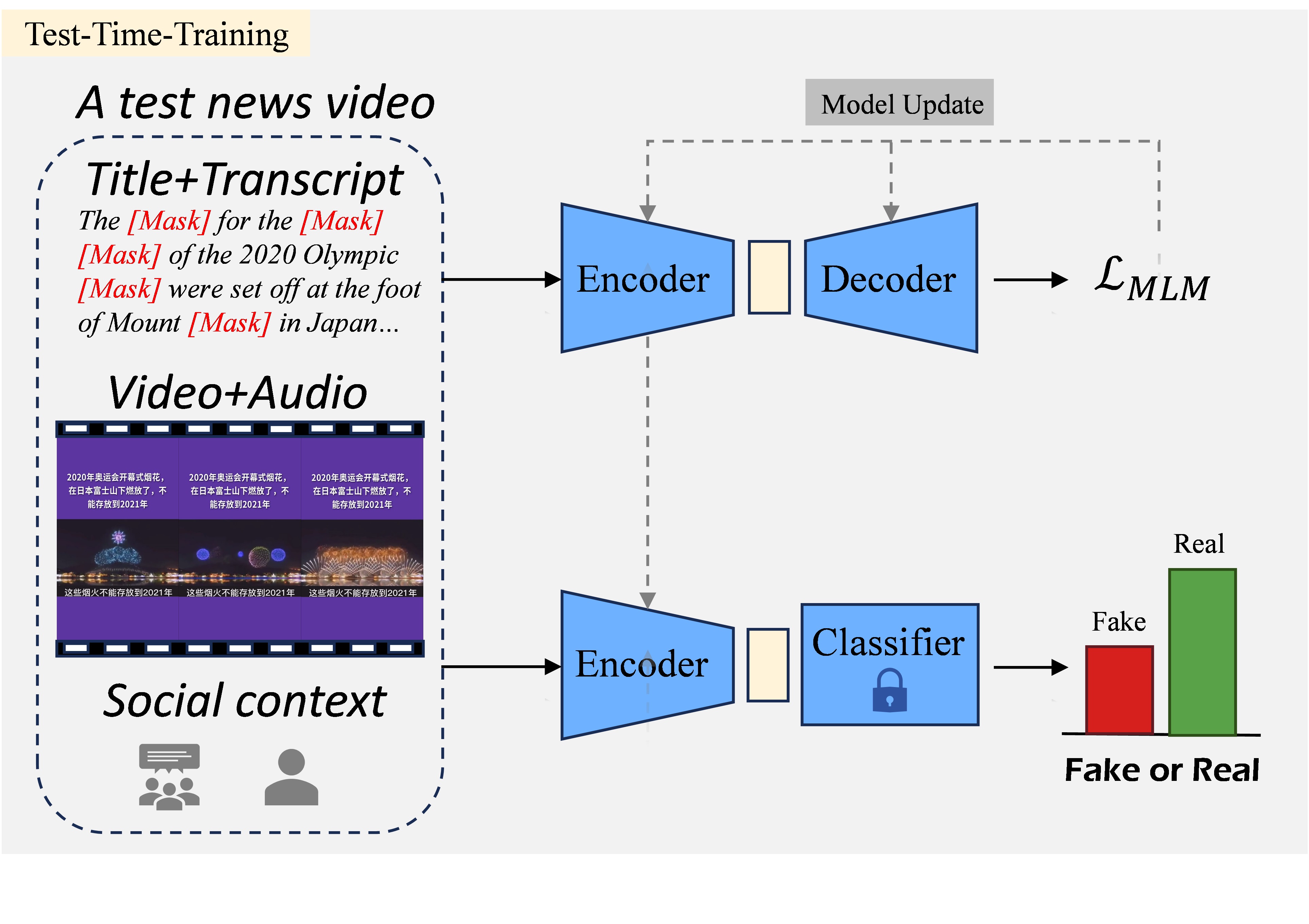}
    \caption{Overview of our testing time training methods. During the testing time training phase, we update the weights of the encoder by using a self-supervised task to adapt to the distribution shift of the test data, while the weights of the classifier are fixed during this process. Then, we use the updated weights to predict the test data.}
    \label{ttt}
\end{wrapfigure}

We introduce a \textbf{T}est-\textbf{T}ime \textbf{T}raining (TTT) framework to alleviate this limitation. With TTT algorithm as the core, we design an auxiliary task based on the self-supervised \textbf{M}ask \textbf{L}anguage \textbf{M}odeling (MLM) to force the model to adapt to the distribution of test data to capture the inherent features hidden in fake news videos. Specifically, we design a multimodal masked Transformer module (M-Transformer$^2$) to integrate information from audio and video respectively to predict the truth value of masked words. In the test-time training phase, we fine-tune the trained model through auxiliary tasks (converted into a self-supervised learning problem), as shown in Fig.~\ref{ttt}, which makes our model focus not only on the binary labels, but also on the distribution of the data itself, ensures the adaptability and robustness of the model in dynamic environment.

Our method does not improve the robustness of the model rely on specific structures, but rather on unlabeled data. The core idea of the TTT framework is that from an information perspective, previously we only used the informational from the training set (supervised learning) to train neural networks, but in fact, the testing set also provides information from the perspective of data distribution. Our research brings hope that in the real world, we only need to collect some news from emergencies (without expensive manual annotation) to fine-tune the model, which can significantly improve its adaptability to emergencies. This focus on adaptive learning marks an significant step forward in combating misinformation videos.

The main contributions of this paper are as follows:
\begin{itemize}
    \item \textbf{Idea:} We for the first time consider the problem of model robustness for detecting fake news on short video platforms. Existing methods are often ineffective when dealing with emergencies due to the lack of appropriate mechanisms to adapt to the dynamic social media environment. It is imperative to transcend current paradigms and improve the adaptability of the model to emergencies.
    \item \textbf{Method:} We propose T$^3$SVFND, a novel fake news videos detection model, which addresses the key challenge by introducing the TTT framework and utilizing a carefully designed self-supervised task based on MLM to learn latent features of multimodal data. To our knowledge, this is the first study on the robustness of fake news videos detection.
    \item \textbf{Effectiveness:} We conducted extensive experiments on a large-scale fake news short videos dataset, compared with the most advanced methods, T$^3$SVFND achieves SOTA results in both event-based and time-based data segmentation scenarios (up to 2.48\% and 3.32\% in accuracy). The ablation study validated the effectiveness of the different modules. Our codes is publicly available in https://github.com/ZhangLiyuan11/TTTSVFND.
\end{itemize}


\section{Related Work}
\label{sec:relatedwork}
In this section, we review the related work on (1) Fake news videos detection; (2) Domain adaptation in fake news detection; (3) Test-time Training Framework. We also explain the innovative aspects of our work.
\subsection{Fake News Videos Detection}

With the development of deep neural networks \cite{ma2023contrastive,ma2023propagation}, some studies have extracted multimodal features of news videos and established cross-modal correlation models, e.g., Choi et al. \cite{choi2021using} attempted to identify differences in stance through differences in topic distribution
between titles/descriptions and comments. Recently, Qi et al. \cite{qi2023fakesv} collected a new dataset of large-scale Chinese fake news short videos and proposed a benchmark by fusing multimodal clues through multiple cross-attention modules. The NEED framework \cite{qi2023two} proposed incorporating neighbor relationship in an event for fake news videos detection. Wu et al. \cite{wu2024interpretable} made the results of fake news videos detection more reasonable by backtracking the decision-making process of the model. 

However, they did not take into account the challenges brought by the dynamic changes in the real world. To overcome the limitations of existing works, we propose a new test-time training based fake news videos detection model T$^3$SVFND, which introduces the TTT framework and a carefully designed MLM task to improve classification performance and reduce the potential risk of feature distribution shift.

\subsection{Domain Adaptation in Fake News Detection}
Many works research the robustness of fake news detection model based on image and text \cite{mosallanezhad2022domain,li2021multi,silva2021embracing}. For example, Mosallanezhad et al. \cite{mosallanezhad2022domain} proposed a domain adaptive detection framework that utilizes reinforcement learning combined with auxiliary information. Li et al. \cite{li2021multi} applied researchers' prior knowledge of fake news to the target domain through weak supervision. Silva et al. \cite{silva2021embracing} combined domain specific knowledge and cross domain knowledge in news records to detect fake news from different domains. However, in the field of fake news videos detection, how to effectively improve model robustness is still an open question, and the application of robustness techniques has not yet been deeply explored.
\subsection{Test-time Training Framework}
Testing-time training is a general framework for handling supervised tasks facing changes in data distribution \cite{sun2020test,liu2021ttt++}. This method includes a self-supervised auxiliary task that not only takes effect during the training phase, but also fine-tunes the model using unlabeled test data during the test-time training phase to adapt to changes in data distribution. In the field of computer vision, there have been many works that apply TTT methods to images \cite{mirza2023mate,wang2020tent,liang2020we,gandelsman2022test,mirza2022norm}, and the experiments of these works show that the framework effectively reduces the performance gap of the model in the face of distribution shift. 

Qian rt al. \cite{zhang2024t3rd} also introduced TTT into the field of fake news detection. However, their method is based on news propagation graphs and introduces Contrastive Learning (CL) as an auxiliary task, including local contrastive learning and global contrastive learning, with the aim of learning graph invariant and node invariant representations on the propagation graph of news. Our method is based on the multimodal content of news videos (text, visual, and audio) to detect fake news videos. We also extend MLM to the multimodal domain, reconstructing text representations by combining information from different modalities, and alleviating the problem of data distribution shift by collaborating with the TTT framework.


\section{METHOD}
\label{sec:method}
\subsection{Model Overview}
\begin{figure}
    \centering
    \includegraphics[width=\textwidth]{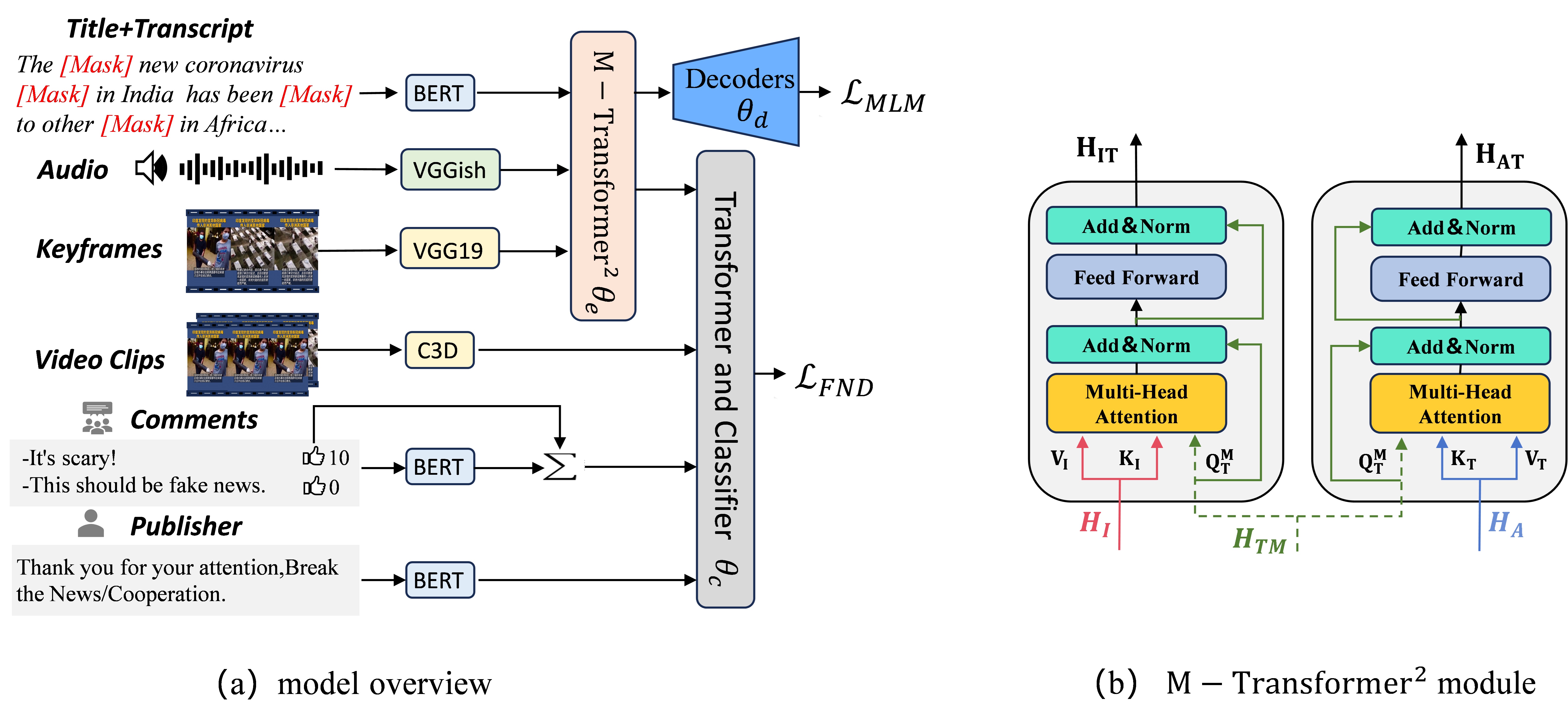}
    \caption{(a) Overview of our entire model architecture. We first perform random masking on the input text. These partially masked text features, along with visual and audio features, are sent to the M-Transformer$^2$ module for cross-modal interaction, and then to the decoders for joint reconstruction to obtain reconstruction loss. The output of the M-Transformer$^2$ module and other features that do not participate in auxiliary tasks are fed into a Transformer for fusion. In order to adapt to the distribution of test samples, we only use reconstruction loss to fine-tune the trained model during the test-time training phase. Finally, evaluate using the updated encoder weights. (b) Architecture of our proposed M-Transformer$^2$ module}
    \label{model}
\end{figure}

The goal of our proposed T$^\text{3}$SVFND framework is to enhance the robustness of the fake news videos detection model. Fig.~\ref{model} (a) illustrates the overall architecture of the T$^\text{3}$SVFND. We define the target for detecting the authenticity of news as $\mathcal{L}_ {FND}$, which takes effect during the training phase. Following the principles of the TTT framework, we introduce self-supervised learning with MLM as the auxiliary task to adapt to data distribution, with the goal defined as $\mathcal{L}_ {MLM}$, which takes effect during the training phase and test-time training phase. All the multimodal features were fed into a Transformer module for feature fusion, and we use a linear layer as the classifier.



\subsection{Feature Extraction}
The input of our model is a multimodal news video, which contain video, title, comments, and publisher introduction. The auxiliary tasks we design mainly involve textual, audio, and static visual features. In order to achieve fair comparison, we simultaneously considered the social context and video clip features contained in news video samples and followed the setting of the benchmark \cite{qi2023fakesv}.
\textbf{Textual Encoder.}
For news videos, the length of the title is usually limited, and a large amount of text information is embedded in the video. Therefore, we use ffmpeg\footnote{https://ffmpeg.org/} and PaddleOCR\footnote{https://github.com/PaddlePaddle/PaddleOCR} tools to extract video transcription from video and concatenate it with the title. For feature extraction, We use pre-trained BERT \cite{cui2021pre} as the textual encoder, the composed text is further fed into the BERT to extract the token-level textual semantic features $H_T=[t_1,..., t_l]$, where $t_i$ represents the feature of the $i\mbox{-}th$ word, and $l$ is the length of the text.

\textbf{Visual Encoder.} For visual feature extraction, considering that the main semantics of a video can be summarized in keyframes. We use a pre-trained VGG19 \cite{liu2021ttt++} model as visual encoder to extract static frame-level visual features $H_I=[i_1,..., i_m]$, where $m$ is the number of keyframes. We also use the pre-trained C3D model to extract the motion features $H_V = [v_1, ..., v_m]$, an average operation is applied to obtain the aggregated motion feature $x_V$.

\textbf{Audio Encoder.} Audio is also one of the main modalities of news videos, which not only contains semantic features, but also unique information such as environmental sounds and background music. For audio feature extraction, We use a pre-trained VGGish model \cite{hershey2017cnn} for extracting frame-level features $H_A=[a_1,..., a_n]$ from separated audio files, where $n$ is the number of audio frames.

\textbf{Comments.} We use the pre-trained BERT to extract the features of each comment, that is $H_C = [c1, ..., ck]$, and use the number of likes for weighted aggregation to obtain the comments vector $x_C$.

\textbf{Publisher.} We feed the introduction of the publisher into the pre-trained BERT. The embedding of the [CLS] token is extracted as the user features $x_P$.


\subsection{The Auxiliary Task Design}
\subsubsection{Language Mask.} The MLM task is a well-known text auto-encoder pre-training task, the self-supervised nature allows it to serve as an auxiliary task for TTT. We proportionally replace the token representations of these texts with MASK tags using random masking as input to the co-attention module. The masked text sequence is denoted as $H_{TM}$ and the mask ratio $m$ is a hyper-parameter. Then, the extracted visual, audio features and the masked text features are further processed through a multimodal masked Transformer network (M-Transformer$^2$).
\subsubsection{Masked Transformer Design.} The effectiveness of multimodal learning has long been proven \cite{salvi2024multi,huang2021makes}, co-attention based on Transformer architecture has been widely applied in cross-modal interaction \cite{xu2023multimodal,wu2021multimodal,qian2021hierarchical,liang2021multibench}. This network aligns fine-grained features from different modalities through cross-attention, effectively integrating multimodal features while filtering out noise or irrelevant context across modalities. In order to adapt to the MLM based auxiliary task we designed, we developed a multimodal masked Transformer network (M-Transformer$^2$). As shown in Fig.~\ref{model} (b), the M-Transformer$^2$ module consists of two parallel multimodal Transformer units, aim to align and capture complementary information existing in different modalities. We use the masked text features as the query $Q_M$, and the audio and visual features as keys $K_A,K_I$ and values $V_A,V_I$ respectively. In each Transformer unit, the dependency relationship between the two modalities of the inputs is captured through a multi-head cross-attention mechanism. Taking the interaction process between audio sequence and masked text sequence as an example, we generate text features with enhanced audio features in the form of:
\begin{align}
    H_{AT}=({||}^N_{n=1}softmax(\frac{Q_MK_A^T}{\sqrt d})V_A)W_{AT}
\end{align}
 where $N$ is the number of heads and $W_{AT}\in R^{d\times d}$ represents the linear transformation of the output. These head features are re summarized through a linear layer. After the co-attention module, the important news elements in the fine-grained features of the single modality are highlighted and supplemented by others.
\subsubsection{Joint Reconstruction.} We reconstruct the original signal from the masked input of text under the condition of non masked input in another modalities. Specifically, the auxiliary task is divided into two parts with the same operation, the keyframes sequence $H_I$ and audio sequence $H_A$ are respectively used to reconstruct the original text sequence $H_T$, along with the masked text sequence $H_{TM}$. 
Taking the reconstruction process based on audio and masked text features as an example, $H_A$ and $H_{TM}$ are simultaneously fed into the audio-text co-attention encoder to obtain the masked text sequence $H_{AT}$ enhanced by the audio features. $H_{AT}$ is then fed into the audio-text cross-modal decoder $g^{de}_{at}$ and predicts the truth value of the masked token in the text sequence. Similarly, we denote the masked text sequence enhanced by visual features as $H_{IT}$, and the visual-text cross-mode decoder as $g^ {de}_{it}$. The two decoders we use both consist of a Transformer block and a fully connected layer. We use cross entropy as the loss function to evaluate the MLM task, and the loss of the entire MLM task is defined as:
\begin{align}
    \mathcal{L}_{M L M}=\mathcal{L}_{C E}\left(y_{T}^{M}, \hat{y}_{A T}^{M}\right)+\mathcal{L}_{C E}\left(y_{T}^{M}, \hat{y}_{I T}^{M}\right)
\end{align}
among them,
\begin{align}
    \hat{y}_{A T}^{M}=g_{a t}^{d e}\left(\text { Transformer }\left(H_{A}, H_{T M}\right)\right) \\
\hat{y}_{I T}^{M}=g_{i t}^{d e}\left(\text { Transformer }\left(H_{I}, H_{T M}\right)\right)
\end{align}
where the superscript $M$ is the data corresponding to the masked signal, while $y_T^M$ is the truth value of the masked text token, and $\mathcal{L}_{CE}$ represents cross entropy. According to our setup, the model must reconstruct the masked text token by focusing on audio or video features, which can force the model to learn the interactions between cross-modal features while adapting to different data distributions.

 \subsection{Feature Fusion and Classification}
For the final fusion, we first average $H_{AT}$, $H_{IT}$ and then obtain the features $x_{AT}$, $x_{IT}$. Till now, we have obtained five features related to fake news videos, including cross-modal features $x_{AT}$ and $x_{IT}$, video motion feature $x_V$, comment feature $x_C$, and publisher feature $x_P$. We use self-attention to model the correlation between different features. Specifically, we concatenate five features into a sequence and feed it into a standard Transformer layer, applying average pooling to obtain the final feature of fake news videos.

We use a classifier with a fully connected layer and softmax activation. The goal of each news video is to minimize the binary cross-entropy loss function as follows:
\begin{align}
    \mathcal{L}_{FND}=-[(1-y)log\hat{y}+(1-y)log(1-\hat{y})]
\end{align}
where $y$ denotes the ground-truth label and $\hat{y}$ is the prediction result of the classifier.

\subsection{The Test-time Training Framework}
\subsubsection{Training Phase.} During the training phase, our model minimizes the weighted sum of the supervised loss $\mathcal{L}_{FND}$ for fake news videos detection and the auxiliary loss $\mathcal{L}_{MLM}$ based on MLM on the training set, with weight $\alpha$ being an adjustable hyper-parameter. And update all learnable weights simultaneously:
\begin{align}\label{para}
    min\ \ \mathcal{L}_{train}(\theta_e,\theta_c,\theta_d)=\mathcal{L}_{FND}+\alpha\mathcal{L}_{MLM}
\end{align}
\subsubsection{Test-time Training Phase.} During the test-time-training phase, we fixed all parameters except for M-Transformer$^2$ and decoder, minimized the auxiliary self-supervised task on the test set to fine-tune the pre-trained model, and $\theta_e^{*}$ and $\theta_d^{*}$ are the weights of the trained M-Transformer$^2$ and decoder:
\begin{align}
    min\ \ \mathcal{L}_{MLM}(\theta_e^*,\theta_d^*)
\end{align}
\subsubsection{Testing Phase.} In the testing phase, we fix all the weights of the model and predict the labels based on the optimal parameters obtained.
\begin{align}
    \hat y=f(\theta_e^{*'},\theta_c^{*'})
\end{align}
where $x$ is the test sample. Refer to Algorithm \ref{alg:algorithm1} for a detailed explanation of the training process.  

\begin{algorithm}[t]
    \caption{Model training of T$^\text{3}$SVFND}
    \label{alg:algorithm1}
    \renewcommand{\algorithmicrequire}{\textbf{Input:}}
    \renewcommand{\algorithmicensure}{\textbf{Output:}}
    \begin{algorithmic}[1]
        \REQUIRE $D$ for FND.\leftline{Model parameters: $\theta_c$,$\theta_e$,$\theta_d$.}
        Hyperparameters:$\alpha$.Masking ratio m.  
        \ENSURE Model parameters:$\theta_c$ and $\theta_e$.  
        \STATE  Initialize Model parameters;
        \FOR{not converge}
            \STATE \# Tranining
            \STATE Sample minibatch from $D_{train}$;
            \STATE argmin$\mathcal{L}_{Train}=\mathcal{L}_{FND}+\alpha\mathcal{L}_{MLM}$;
            \STATE Update parameters $\theta_c$,$\theta_e$and$\theta_d$ by Adam;
            \STATE \# Test Time Training
            \STATE Sample minibatch from $D_{test}$;
            \STATE argmin$\mathcal{L}_{MLM}$;
            \STATE Update parameters $\theta_e$and$\theta_d$ by Adam;
            \STATE \# Test
            \STATE Sample minibatch from $D_{test}$;
            \STATE predict test data label by $\theta_c$,$\theta_e$;
        \ENDFOR
    \end{algorithmic}
\end{algorithm}

\section{Experiments}



\subsection{Dataset}
We conducted extensive experiments on the FakeSV dataset \cite{qi2023fakesv} to evaluate our proposed method T$^\text{3}$SVFND, which is currently the only large-scale short video fake news dataset that provides rich multimodal clues. FakeSV collects news videos from popular Chinese short video platforms such as Douyin (the equivalent of TikTok in China), and contains 1827 fake news videos and 1827 real news videos. This dataset divides news videos into 738 events and two data split strategies were provided: temporal and event based. For event split, we evaluated them through five-fold cross validation and reported the average based on five runs. For each folding, the dataset is divided into a train set and a test set with a sample ratio of 4:1 at the event level, ensuring that there is no event overlap between different sets and avoiding the model from detecting fake news videos by memorizing event information \cite{wang2018eann}.

\subsection{Baseline Methods}
We compared the proposed model with a range of strong baselines, including handcrafted features-based baselines, neural networks-based baselines, and (multimodal) large language model ((M)LLM) baselines, as follows:

\textbf{LLM Baselines:} (1) \textbf{GPT-4} \cite{openai2024gpt}, one of the most powerful LLMs currently available, used for prediction based on video news titles and extracted screen text. (2) \textbf{GPT-4o}, a variant of GPT-4 that supports visual input, We include the keyframes of the video in the input. (3) \textbf{Video-LLaMA2} \cite{cheng2024videollama}, a multimodal large language model tailored for video understanding, we include the video in the input. For the implementation details of LLM baselines, please refer to the Appendix.


\textbf{Handcraft Feature-based Baselines:} (1) \textbf{HCFC-Hou} \cite{hou2019towards}, used language features for speech and text, acoustic-emotional features, and user-engagement features for classification. (2) \textbf{HCFC-Medina} \cite{serrano2020nlp}, extracted tf-idf vectors from the title and comments, and classified them using a traditional machine learning classifier. 

\textbf{Neural Network-based Baselines:} (1) \textbf{TikTec} \cite{shang2021multimodal}, fused visual and speech information for classification using the co-attention module. (2) \textbf{FANVM} \cite{choi2021using}, modelled the topic distribution differences between titles and comments, extract topic-independent multimodal features for classification. (3) \textbf{SVFEND} \cite{qi2023fakesv}, fused multimodal cues using multiple Transformer modules.

\begin{table*}[t]
\scriptsize
\centering
\caption{Performance comparison of different methods and split rules. In the case of event split, we reported the mean of five folds cross validation.}
\label{tabt1}
\renewcommand{\arraystretch}{1.3}
\begin{adjustbox}{max width=\textwidth}
\begin{tabularx}{\textwidth}{>{\centering\arraybackslash}X >{\centering\arraybackslash}X >{\centering\arraybackslash}X >{\centering\arraybackslash}X >{\centering\arraybackslash}X >{\centering\arraybackslash}X >{\centering\arraybackslash}X >{\centering\arraybackslash}X >{\centering\arraybackslash}X >{\centering\arraybackslash}X >{\centering\arraybackslash}X >{\centering\arraybackslash}X >{\centering\arraybackslash}X}

\toprule
\multicolumn{1}{c}{\multirow{2.5}{*}{Data Split}} & \multicolumn{2}{c}{\multirow{2.5}{*}{Method}} & \multicolumn{2}{c}{\multirow{2.5}{*}{Accuracy}} & \multicolumn{2}{c}{\multirow{2.5}{*}{Macro F1}} & \multicolumn{3}{c}{Real} & \multicolumn{3}{c}{Fake} \\
\cmidrule(l){8-13}
& & & & & & & F1 & Recall & Prec. & F1 & Recall & Prec. \\
\midrule
\multicolumn{1}{c}{\multirow{9}{*}{Temporal} }
& \multicolumn{2}{c}{GPT-4}    & \multicolumn{2}{c}{77.84} & \multicolumn{2}{c}{77.84} & 78.38 & 75.16 & 78.88 & 77.82 & 78.30 & 76.80 \\
& \multicolumn{2}{c}{GPT-4o}   & \multicolumn{2}{c}{70.48} & \multicolumn{2}{c}{70.46} & 72.65 & 68.27 & 71.29 & 70.29 & 67.25 & 69.63 \\
& \multicolumn{2}{c}{VideoLLaMA2} & \multicolumn{2}{c}{58.79} & \multicolumn{2}{c}{58.35} & 47.26 & 63.28 & 54.11 & 70.98 & 55.99 & 62.60 \\
\cdashline{2-13}
& \multicolumn{2}{c}{HCFC-Hou}  & \multicolumn{2}{c}{74.91} & \multicolumn{2}{c}{73.61} & 73.46 & 86.51 & 79.46 & 77.72 & 60.08 & 67.77 \\
& \multicolumn{2}{c}{HCFC-Medina} & \multicolumn{2}{c}{76.38} & \multicolumn{2}{c}{75.83} & 77.50 & 81.58 & 79.49 & 74.77 & 69.75 & 72.17 \\
\cdashline{2-13}
& \multicolumn{2}{c}{TikTec}  & \multicolumn{2}{c}{73.99} & \multicolumn{2}{c}{73.82} & 75.21 & 68.58 & 71.74 & 73.03 & 79.00 & 75.90 \\
& \multicolumn{2}{c}{FANVM}   & \multicolumn{2}{c}{79.70} & \multicolumn{2}{c}{79.49} & 78.99 & 75.81 & 77.37 & 80.26 & 82.99 & 81.61 \\
& \multicolumn{2}{c}{SVFEND}  & \multicolumn{2}{c}{81.18} & \multicolumn{2}{c}{81.11} & \textbf{85.71} & 75.00 & 80.00 & 77.63 & \textbf{87.41} & 82.23 \\
\cdashline{2-13}
& \multicolumn{2}{c}{\textbf{T$^\text{3}$SVFND(Ours)}}    & \multicolumn{2}{c}{$\textbf{84.50}_{(+3.32\%)}$} & \multicolumn{2}{c}{$\textbf{84.24}_{(+3.13\%)}$} & 81.51 & \textbf{82.91} & \textbf{82.20} & \textbf{86.84} & 85.71 & \textbf{86.27} \\

\midrule
\multicolumn{1}{c}{\multirow{9}{*}{Event}}
& \multicolumn{2}{c}{GPT-4}    & \multicolumn{2}{c}{76.84} & \multicolumn{2}{c}{76.84} & 79.67 & 74.44 & 73.93 & 74.00 & 79.24 & 79.75 \\
& \multicolumn{2}{c}{GPT-4o}   & \multicolumn{2}{c}{69.37} & \multicolumn{2}{c}{69.37} & 68.38 & 70.35 & 64.93 & 69.33 & 75.31 & 73.80 \\
& \multicolumn{2}{c}{VideoLLaMA2} & \multicolumn{2}{c}{58.52} & \multicolumn{2}{c}{57.14} & 46.12 & 53.23 & 49.42 & 68.24 & 61.77 & 64.85 \\
\cdashline{2-13}
& \multicolumn{2}{c}{HCFC-Hou}  & \multicolumn{2}{c}{68.94} & \multicolumn{2}{c}{68.53} & 65.52 & 64.41 & 64.96 & 71.62 & 72.60 & 72.11 \\
& \multicolumn{2}{c}{HCFC-Medina} & \multicolumn{2}{c}{70.27} & \multicolumn{2}{c}{69.91} & 67.67 & 65.69 & 66.67 & 72.30 & 74.05 & 73.16 \\
\cdashline{2-13}
& \multicolumn{2}{c}{TikTec}  & \multicolumn{2}{c}{73.25} & \multicolumn{2}{c}{73.09} & 74.79 & 70.68 & 71.06 & 72.04 & 75.49 & 75.13 \\
& \multicolumn{2}{c}{FANVM}   & \multicolumn{2}{c}{74.17} & \multicolumn{2}{c}{74.06} & 76.89 & 71.28 & 72.33 & 72.04 & 76.93 & 75.58 \\
& \multicolumn{2}{c}{SVFEND}  & \multicolumn{2}{c}{78.45} & \multicolumn{2}{c}{78.41} & 79.79 & \textbf{77.80} & 78.69 & 77.07 & 79.46 & 78.13 \\
\cdashline{2-13}
& \multicolumn{2}{c}{\textbf{T$^\text{3}$SVFND(Ours)}}    & \multicolumn{2}{c}{$\textbf{80.93}_{(+2.48\%)}$} & \multicolumn{2}{c}{$\textbf{80.81}_{(+2.40\%)}$} & \textbf{81.09} & 76.89 & \textbf{78.94} & \textbf{80.92} & \textbf{82.54} & \textbf{82.69} \\

\bottomrule
\end{tabularx}
\end{adjustbox}
\end{table*}

\subsection{Performance Comparison}
Table \ref{tabt1} shows the performance comparison between T$^\text{3}$SVFND and other methods, the results show that the proposed method outperforms all compared methods in terms of accuracy scores for each partition, demonstrating the effectiveness of our proposed model. Specifically, in the case of event split and temporal split, T$^\text{3}$SVFND is 2.48\% and 3.32\% higher than the previous best data respectively.

The zero-shot LLM-based methods, exhibit stable performance in different data split scenarios, but even the most powerful LLM is far inferior to the latest models specifically tailored for fake news videos detection, indicating the necessity of designing specialized models. Notably, the performance of MLLM with visual input has decreased instead, indicating that large models still have shortcomings in complex visual understanding. Neural network-based baselines generally outperform handcraft feature-based baselines, demonstrating the superiority of neural network models in handling complex fake news videos detection task. All of these multimodal baselines showed significant performance degradation in the event split scenario, T$^\text{3}$SVFND optimized this by introducing the TTT training framework and MLM auxiliary tasks. In the event split scenario, T$^\text{3}$SVFND performed better.

In the case of event split, the performance of T$^\text{3}$SVFND is lower than that of temporal split, possibly due to the existence of long-standing fake news events. In the real world, we can only train detectors through historical data, therefore dataset based on temporal split are more in line with the real situation in the real world. The event split method better reflects the generalization performance of the model in the face of unexpected events that have never been seen before. Compared with the previous model, T$^\text{3}$SVFND has a higher performance in event split, demonstrating its superiority.






\subsection{Ablation Studies}
To research the effectiveness of each component in T$^\text{3}$SVFND, we conduct extensive ablation studies. We simplified the model as follows: \textbf{(1) w/o TTT.} We remove the test-time training algorithm framework and replace it with a traditional framework, with the auxiliary task only taking effect during the training phase. \textbf{(2) w/o MLM.} We remove mask language model (MLM) task and use traditional training framework (auxiliary tasks are also not effective). \textbf{(3) w/o Tran.} We remove multimodal Transformer as the fusion module and only use unmasked text features for reconstruction, without combining other modalities. \textbf{(4) w/o V.} We remove the keyframe faetures and there related parts from the news, and only use audio features and masked text features for joint-reconstruction. \textbf{(5) w/o A.} We remove audio features and there related parts from the news, and only use keyframe features and masked text features for joint-reconstruction.

The results of which are detailed in Table \ref{tabt2}. We firstly focus on the core of T$^\text{3}$SVFND: TTT and MLM. It can be observed that introducing the TTT training framework and MLM based auxiliary tasks can effectively improve the model's detection performance in the face of emergency news. We also explored removing Transformer as a fusion module and only using unmasked text features for reconstruction. This variant exhibits relatively low performance, indicating that the M-Transformer$^2$ module can not only perform auxiliary tasks well, but also enable effective cross-modal interaction, thereby utilizing the complementarity of multimodal features.

Further exploration of each specific aspect in our designed auxiliary task: joint reconstruction with video and audio features separately. By systematically removing each aspect, the results confirmed that joint reconstructions of masked text using both visual and audio features have made beneficial contributions to the robustness of the model, which underscores the synergy that their integration brings to the effectiveness of T$^\text{3}$SVFND in detecting fake news videos.

\begin{table}[t]
  \renewcommand{\arraystretch}{1.2}
  \setlength{\tabcolsep}{2mm}
\begin{center}
\caption{Ablation study on different modalities under event split. The standard deviation values are ignored for simplicity.}
\label{tabt2}
\begin{tabular}{ccccc|cccc} 
\midrule
\multicolumn{5}{c|}{\multirow{1}{*}{Module}} & \multicolumn{4}{c}{\multirow{1}{*}{Result}}\\
\midrule
TTT &  MLM & Trans & V & A & Acc. & F1 & Prec. & Recall\\
\midrule
\ding{52} &  \ding{52} & \ding{52} & \ding{52} & \ding{52} & \textbf{80.93} & \textbf{80.81} & \textbf{79.91} & \textbf{80.25}\\
\cdashline{1-7}[3pt/5pt]
 &  \ding{52} & \ding{52} & \ding{52} & \ding{52} & 77.58 & 77.32 & 76.58 & 77.12\\
 &  & \ding{52} & \ding{52} & \ding{52} & 77.53 & 77.28 & 76.94 & 77.48\\
\cdashline{1-7}[3pt/5pt]
\ding{52} &  \ding{52} &  &  &  & 78.49 & 78.45 & 77.59 & 78.25\\
\ding{52} &  \ding{52} & \ding{52} & \ding{52} &  & 79.22 & 79.16 & 78.92 & 79.06\\
\ding{52} &  \ding{52} & \ding{52} &  & \ding{52} & 79.50 & 79.41 & 78.27 & 78.80\\
\midrule
\end{tabular}
\end{center}
\end{table}

\subsection{Hyper-parameter Research}
 \subsubsection{Hyper-parameter \textbf{$\alpha$}.} Study the impact of setting hyper-parameter $\alpha$ in \eqref{para} on performance (accuracy). As shown in Fig.~\ref{visualization} (a), with the increase of $\alpha$, the performance of the model gradually improves and reaches its peak at parameter 1. However, there is difference between training events and testing events varies under different data split scenarios, which may lead to certain biases.
 
 \subsubsection{Hyper-parameter \textbf{$m$}.} 
\begin{figure}
    \centering
    \includegraphics[width=\textwidth]{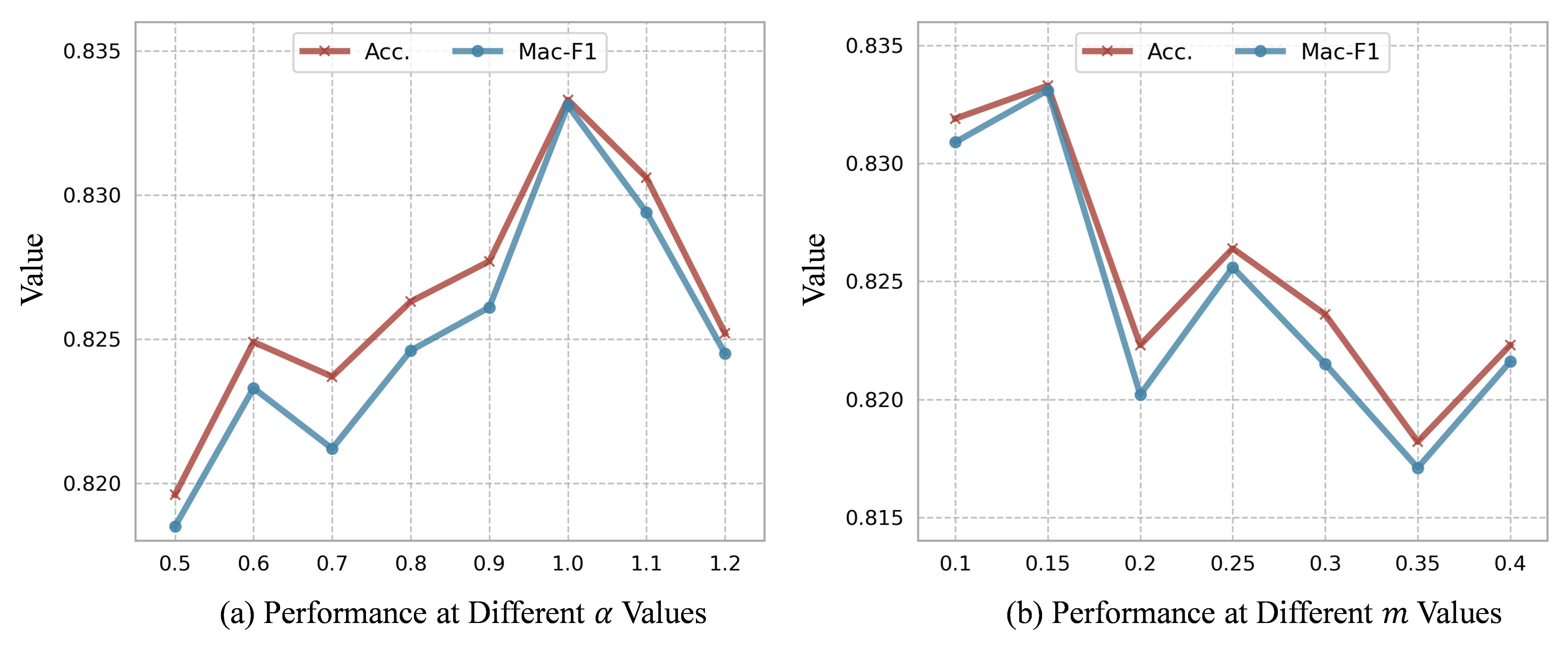}
    \caption{Visualization results of hyper-parameters analysis. In the event set numbered 4 as the testing set, we first fix m to 0.15 to explore the impact of changes in $\alpha$ on performance.}
    \label{visualization}
\end{figure}

In our method, for each input text sequence, we mask a portion of the words based on the masking rate $m$, which is a predefined hyper-parameter. We fix $\alpha$ to 1 and investigate the impact of setting the mask ratio on performance. As shown on the right of Fig.~\ref{visualization}, the best results were achieved when the mask ratio was 0.15. This behavior is similar to BERT (whose typical masking ratio is 15\%). 
 Due to the high semantics the masked language features possessed, excessive high masking ratio may lead to unstable model training and increase the risk of overfitting.
 

\section{Conclusion and Future Work}
In this work, we propose T$^\text{3}$SVFND, a new method for multimodal fake news videos detection. 
Introducing the TTT training framework and MLM as auxiliary tasks , which can effectively enhance the ability of fake news videos detection models to adapt to changes in data distribution in the real world. 
The experimental results demonstrate the effectiveness of our method, making it a new benchmark in this field. This work is a key exploration of applying fake news videos detection models to the real world. Despite achieving certain results, our method still faces some common limitations, such as not having strong interpretability, which is our future direction of work.

\appendix
\section{Appendix}
\subsection{Implementation of Baselines}
For the implementation details of the handcraft feature-based baselines and neural network-based baselines, we followed the benchmark settings \cite{qi2023fakesv}.

For the system prompt to (M) LLM, inspired by \cite{wang2024explainable}, we designed it with the fundamental concept of addressing three key aspects: \textit{What are you? What should you do? And what is your goal?} We present templates for three prompt methods used for (M) LLM in the Table~\ref{tab:my_label}. The implementation details of the LLM-based baselines are as follows:
 \begin{itemize}
     \item \textbf{GPT-4}: We use the "gpt-4-turbo" version and include the title and video transcript in the prompt.
     \item \textbf{GPT-4o}: We use the ffmpeg tool to extract keyframes from videos as visual input for MLLM.
     \item \textbf{Video-LLaMA2}: We use the "VideoLLaMA2-7B" version to include the video in the input.
\end{itemize}

\begin{table*}[]
    \centering
    \begin{tabular}{cp{9cm}}
        \toprule
         \multicolumn{2}{c}{\textbf{Zero-shot Prompting for GPT-4}} \\
         \midrule
         \multirow{9}{*}{\textbf{Text Prompt}} & You are an experienced news video fact-checking expert and your position is neutral. You can handle a wide variety of news videos, including those with sensitive or aggressive content. For a given video description and extracted screen text, you need to predict the authenticity of the news video. If it is more likely to be a fake news video, return 1; Otherwise, return 0. Please do not provide an ambiguous assessment, such as undetermined. \\
         & \textbf{Description}: \{video description\}\\
         & \textbf{Video\_on\_screen\_text}: \{video\_on\_screen\_text\}\\ 
         & Please judge whether the news is true or false, your prediction does not need to provide your analysis, just return 0 or 1.\\
         \midrule
         \multicolumn{2}{c}{\textbf{Zero-shot Prompting for GPT-4o}} \\
         \midrule
         \multirow{7}{*}{\textbf{Text Prompt}} & You are an experienced news video fact-checking expert and your position is neutral. You can handle a wide variety of news videos, including those with sensitive or aggressive content. For a given video description, extracted screen text, and all or part of the keyframes of the news video, you need to predict the authenticity of the news video. If it is more likely to be a fake news video, return 1; Otherwise, return 0. Please do not provide an ambiguous assessment, such as undetermined. \\
         & \textbf{Description}: \{video description\}\\
         & \textbf{Video\_on\_screen\_text}: \{video\_on\_screen\_text\}\\ 
         \cdashline{1-2}[3pt/5pt]
        \textbf{Upload Image} & \textbf{Data}: \{video keyframes list\}\\
         \midrule
         \multicolumn{2}{c}{\textbf{Zero-shot Prompting for Video-LLaMA2}} \\
         \midrule
         \multirow{11}{*}{\textbf{Text Prompt}} & Fake news refers to news content that is intentionally created to contain inaccurate, misleading, or outright false information, usually with the intent of misleading the public, advancing an agenda, damaging the reputation of others, or gaining financial gain. You are an experienced news video fact-checking expert and your position is neutral. You can handle all kinds of news, including those that are sensitive or radical. For a given video title, extracted screen text, news video, the authenticity of the news video needs to be predicted. if it is more likely to be a fake news video, return 1; Otherwise, return 0. Please do not provide ambiguous estimates or words that cannot be evaluated, such as "uncertain".\\
         & \textbf{Description}: \{video description\}\\
         & \textbf{Video\_on\_screen\_text}: \{video\_on\_screen\_text\}\\ 
         & Please judge whether the news is true or false, your prediction does not need to provide your analysis, just return 0 or 1.\\
         \cdashline{1-2}[3pt/5pt]
         \textbf{Upload Video} & \textbf{Data}: \{video\}\\
         \bottomrule
    \end{tabular}
    \caption{The system prompt to (M) LLMs.}
    \label{tab:my_label}
\end{table*}

\bibliographystyle{splncs04}
\bibliography{main.bib}

\end{document}